\documentclass{article}

\usepackage{arxiv}

\usepackage[utf8]{inputenc} % allow utf-8 input
\usepackage[T1]{fontenc}    % use 8-bit T1 fonts
\usepackage{hyperref}       % hyperlinks
\usepackage{url}            % simple URL typesetting
\usepackage{booktabs}       % professional-quality tables
\usepackage{amsfonts}       % blackboard math symbols
\usepackage{nicefrac}       % compact symbols for 1/2, etc.
\usepackage{microtype}      % microtypography
\usepackage{lipsum}
\usepackage{graphicx}
\graphicspath{ {./images/} }
\usepackage{amsmath}
\usepackage{booktabs}
\usepackage{graphicx}
\usepackage{subcaption}

\title{Integrating Genomics into Multimodal EHR Foundation Models}

\author{
 Verily Life Sciences\\
 Jonathan Amar, Edward Liu, Alessandra Breschi, Liangliang Zhang, Pouya Kheradpour,\\
 Sylvia Li, Lisa Soleymani Lehmann, Alessandro Giulianelli, Matt Edwards, Yugang Jia \\
 \texttt{@verily.com} \\
  \And
 Nvidia \\
  David Nola, Raghav Mani, Pankaj Vats, Jesse Tetreault, T.J. Chen\\
  \texttt{@nvidia.com} \\
  \And
   Google\\
 Cory Y. McLean \\
\texttt{cym@google.com} \\
}

\begin{document}
\maketitle
\begin{abstract}
This paper introduces an innovative Electronic Health Record (EHR) foundation model that integrates Polygenic Risk Scores (PRS) as a foundational data modality, moving beyond traditional EHR-only approaches to build more holistic health profiles. Leveraging the extensive and diverse data from the All of Us (AoU) Research Program, this multimodal framework aims to learn complex relationships between clinical data and genetic predispositions. The methodology extends advancements in generative AI to the EHR foundation model space, enhancing predictive capabilities and interpretability. Evaluation on AoU data demonstrates the model's predictive value for the onset of various conditions, particularly Type 2 Diabetes (T2D), and illustrates the interplay between PRS and EHR data. The work also explores transfer learning for custom classification tasks, showcasing the architecture's versatility and efficiency. This approach is pivotal for unlocking new insights into disease prediction, proactive health management, risk stratification, and personalized treatment strategies, laying the groundwork for more personalized, equitable, and actionable real-world evidence generation in healthcare.

\end{abstract}

% keywords can be removed

\begin{figure}
    \centering
    \includegraphics[width=1.\linewidth]{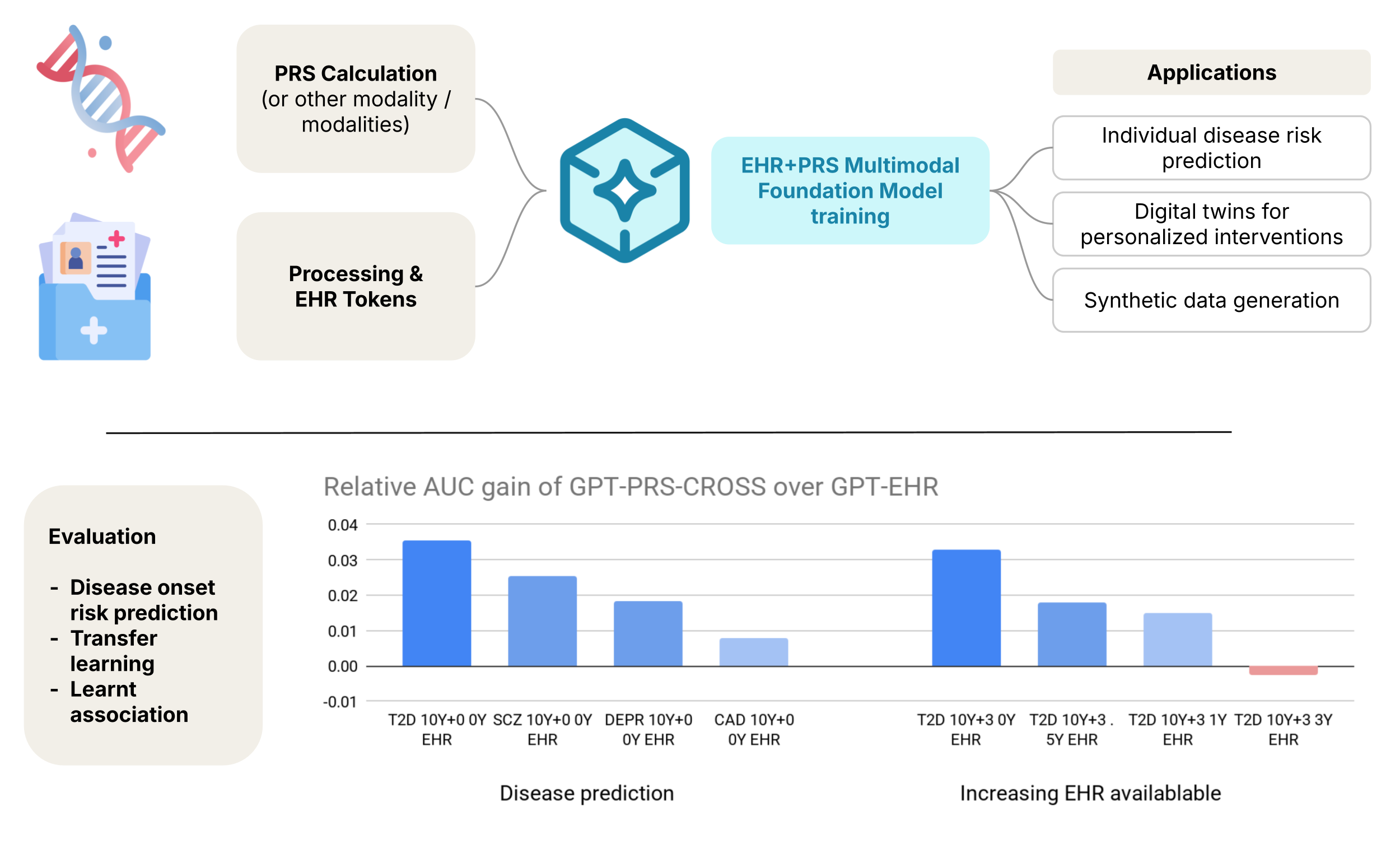}
    \caption{Summary of the Electronic Health Records (EHR) foundational model with multi-modality genomics using polygenic risk scores (PRS).
    \textbf{(Top)} Model is trained using generic next token prediction on the All of Us dataset.
    \textbf{(Bottom)} The FM is evaluated on held-out data for various conditions - Type 2 Diabetes (T2D), Schizophrenia (SCZ), depression (DEPR), and Coronary Artery Disease (CAD) - and medical history settings (increasing EHR made available). AUCs compare the multi-modal EHR+PRS (GPT-PRS-CROSS) against the EHR (GPT-EHR) only foundation model.}
    \label{fig:summary_chart}
\end{figure}

\section{Introduction}
\subsection{General introduction and key idea}
% High-quality, personalized healthcare is built on delivering the right intervention to the right individual at the right time. Achieving this requires more than just evidence-based practice – it depends on consensus-driven guidelines, integrated data, and insights that reflect the full diversity of real-world care. At its core, effective medical care rests on four essential pillars: accurate diagnosis, reliable prognosis, individualized treatment planning, and streamlined clinical workflow. Excelling in each of these areas means drawing on an individual’s longitudinal health history, addressing both diagnostic and prognostic uncertainty, incorporating personal values and goals, and tailoring clinical reasoning to the specific temporal data and care context.
The central goal of modern, high-quality personalized healthcare is to provide therapies and support that are precisely tailored to the unique needs of each person. Reaching this standard involves moving beyond conventional evidence-based practice. It requires leveraging integrated, comprehensive data sources that capture the true diversity of patient experiences and outcomes. Fundamentally, effective medical care hinges on success in several key areas: accurate diagnostics (correctly identifying a condition), reliable prognostics (forecasting a patient's future health), personalized treatment strategies (designing a care plan for the individual), efficient clinical logistics (streamlining the process of care delivery).
Achieving excellence in these domains is not possible without a deep, holistic understanding of a patient's entire medical journey over time. This includes skillfully managing clinical uncertainties, respecting the patient's individual goals and values, and adapting all reasoning to their specific, unfolding circumstances \cite{waxler2025generative}.

\subsection{Overview of Foundational Models (FM) in Electronic Health Records (EHR)}
Foundational models (FMs) trained on longitudinal electronic health records (EHR) are transforming personalized medicine by converting real-world data into actionable insights. Models like ETHOS \cite{renc2024zero}, CLMBR \cite{steinberg2021language}, Foresight \cite{kraljevic2024foresight}, Delphi \cite{shmatko2025learning}, and Curiosity \cite{waxler2025generative} have demonstrated the power of leveraging EHR event streams to predict clinical outcomes for diagnostic and prognostic applications. These models are trained on a generic next-token prediction task with individual health trajectories and can surprisingly perform various clinical and operational prediction tasks without further fine-tuning. By analyzing AI-generated health trajectories in the form of EHR tokens, these models can be used for many risk assessment and predictive tasks \cite{renc2025foundation}, providing sequences that experts can interpret to identify specific risks.

However, these models are often limited by their reliance on a single data modality. While EHRs are widely accessible, they may lack the depth of other data types crucial for personalized medicine; additional data modalities, such as genomics, clinical notes, imaging, surveys, and social determinants of health (SDOH), enable a more holistic understanding of individual health dynamics. However, encoding multiple modalities in one model is difficult due to the inflexibility of how these current models encode data.

In this work, we extend the capabilities of traditional EHR foundational models by introducing a principled framework for integrating additional data modalities into EHR-based foundation models through two proposed architectures:
\begin{itemize}
\item \textbf{Cross-attention}: 
Inspired by encoder–decoder paradigms, we employ cross-attention\cite{lin2022cat} mechanisms to align and integrate representation from diverse  modalities with tokenized clinical histories. 
\item \textbf{Adapter modules}: We insert vectorized modality embeddings within health trajectories, generated via learned adapter\cite{liu2023visual} modules: static modalities (e.g., genomics) are prepended to the tokenized history while dynamic modalities (e.g., survey-based SDOH signals) could be inserted mid-sequence, thereby preserving temporal context while enabling joint modeling.
\end{itemize}

By adapting these methods to integrate diverse data specific to an individual into EHR FMs, we aim to enable more personalized insights and relevant predictions.

\subsection{Using PRS for various risk assessments}
Polygenic risk scores (PRS) represent a powerful tool in personalized medicine, aggregating the effects of multiple genetic variants across an individual's genome to estimate their predisposition to complex diseases. By quantifying the cumulative genetic liability, PRS offer a unique lens through which to assess individual risk for a wide array of conditions, from common chronic illnesses like type 2 Diabetes~\cite{Smith2024_T2D} and cardiovascular diseases~\cite{Norland2024_CHD}, to neuro-degenerative disorders~\cite{Bellou2025_AlzPRS} and certain cancers~\cite{Tamlander2024_CRC}. Integrating PRS into clinical practice allows for earlier identification of at-risk individuals, potentially enabling proactive interventions such as targeted screening, lifestyle modifications, and tailored preventive strategies.
This genetic-informed risk stratification adds to traditional risk factors, providing a more comprehensive view of an individual's health trajectory and long-term outcomes.

In parallel, recent advancements in text-based Large Language Models (LLMs) have demonstrated the power of multimodal approaches in healthcare reasoning. For instance, models like Med Palm\cite{tu2024towards} and Med-Gemini\cite{yang2024advancing} have effectively incorporated various modalities, including PRS and medical images, to enhance diagnostic accuracy and personalized treatment recommendations within text based LLMs. Similarly, Google's PH-LLM\cite{khasentino2025personal} has showcased the utility of wearable data, leveraging adapter-based architectures to seamlessly integrate continuous physiological signals into the learning process. AMIE incorporated multimodality\cite{saab2025advancing} for diagnostic applications.
In the realm of EHR models, Delphi \cite{shmatko2025learning} has also explored the inclusion of PRS, albeit in an indirect manner using ensemble models ex-post, rather than a systematic approach by design. This highlights the ongoing evolution of how genetic information (and various modalities) is incorporated into healthcare foundation models. These examples collectively underscore the growing recognition that a holistic view of members' data, encompassing genetic, imaging, and real-time physiological metrics, is crucial for unlocking the full potential of AI in personalized medicine.

\subsection{Contributions and data}
Our approach is applied to data from the All of Us (AoU) Research Program \cite{all2019all, ramirez2021progress}, covering participants' data in a non-acute care setting. There are 500,000+ participants who have completed the initial steps of the program, as of the V8 cut-off date of Oct 2023.
The All of Us Research Program, a landmark initiative launched by the U.S. National Institutes of Health (NIH), is designed to build one of the most diverse and comprehensive health databases in history. Unlike traditional research cohorts, AoU places a deliberate emphasis on recruiting individuals from historically underrepresented populations in biomedical research, to best reflect the rich diversity of the nation. Notably, the AoU dataset is distinctive not only in its breadth, but also in its richness: each participant's record includes structured clinical events, genomic profiles, wearable data, and survey-derived indicators of disease severity and healthcare access. This enables us to evaluate how the addition of genomic PRS data improves foundation model performance across both operational and clinical targets.

To the best of our knowledge, this research represents a pioneering effort in the study of EHR foundation models utilizing the AoU dataset. This novel application of AoU data is particularly significant as it holds the potential to substantially enhance diversity modeling within EHR FM, leading to more robust and generalizable insights across diverse participant populations.

Additionally, our evaluation conclusively demonstrates that the multimodal integration approach, specifically utilizing PRS improves performance across real-world clinical tasks. This robust enhancement highlights the critical value of combining data modalities into EHR FM. From a clinical perspective, the strategic integration of novel modalities yields significant benefits for patients, including the capability for earlier and more accurate disease predictions, superior disease risk stratification, and a more nuanced understanding of patient vulnerability. This is particularly advantageous for predicting outcomes that are less comprehensively captured when relying solely on structured EHR data.

In summary, our work contributes:
\begin{enumerate}
    \item Architectural designs for multimodal EHR foundation models, employing cross-attention or adapter-based vectorized embeddings to integrate static and dynamic modalities into state of the art decoders.
    \item A novel mechanism that more accurately assesses probability of disease incidence from generative EHR models.
    \item Application of this architecture on the diverse and multimodal All of Us cohort, leveraging genomics through PRS as an additional modality beyond EHR only foundation models.
    \item Qualitative validation of the methods, by manually inspecting learned representations of the participants' data, as encoded by the trained model. The model, despite being trained in a self-supervised fashion, yields interpretable properties by serendipity, giving hope to further utilization of multimodal EHR FM.
    \item Evaluation of multimodal EHR FMs, specifically utilizing PRS broadly, compared to EHR-only FM. We demonstrated that integration of genomics data meaningfully improves performance across multiple real-world clinical and operational tasks, thereby advancing the potential clinical value and scalability of EHR foundation models in healthcare.
\end{enumerate}

By enabling richer, more context-aware modeling of health histories across modalities, our framework lays the groundwork for more personalized, equitable, and actionable real-world evidence generation – from risk prediction to operational resource planning. A summary of our contribution and key results is shown in Figure \ref{fig:summary_chart}.

\section{Methods}
\subsection{Data}
In this study, we applied our methodology to the All of Us dataset.\footnote{
\href{https://www.researchallofus.org/data-tools/data-snapshots/}{https://www.researchallofus.org/}}
The All of Us Research Program is a U.S. nationwide initiative designed to create a diverse, longitudinal health dataset with broad multimodal coverage. Unlike traditional hospital-centric EHR cohorts, All of Us recruits participants from community settings and health provider organizations, with a deliberate emphasis on historically underrepresented populations. The dataset currently encompasses over 800,000 participants and integrates multiple complementary modalities: structured and unstructured EHR data harmonized into the Observational Medical Outcomes Partnership (OMOP) Common Data Model (CDM);\footnote{\href{https://www.ohdsi.org/data-standardization/}{https://www.ohdsi.org/data-standardization/}}
survey responses capturing health behaviors, social determinants of health, and healthcare access; biospecimens enabling genomic and molecular profiling; and optional digital health data from wearable devices. This dataset provides a uniquely rich substrate for training multimodal foundation models of EHR data, enabling architectures that jointly represent clinical events, genomic risk factors, and social context. In particular, AoU is a powerful resource for developing and evaluating multimodal models that go beyond conventional EHR FM towards more holistic representations of individual health.

\paragraph{Data pre-processing}
We extracted clinical records from the OMOP common data model, which includes conditions, procedures, laboratory results and measurements, visit information, and basic demographics (i.e., sex, gender, and race). This data was processed into a Medical Event Data Standard \footnote{ \href{https://medical-event-data-standard.github.io/docs/intro_pages/what_is_MEDS}{https://medical-event-data-standard.github.io/docs/}}
(MEDS) like format, which allows the model-oriented representation of medical events timestamps and is conveniently used in EHR FM. These “medical events” capture individual data points from OMOP format.
To ensure data quality and account for technical model constraints and computational feasibility  considerations of FM model training, participants with fewer than 100 or more than 2000 measurements (i.e., rows of the MEDS-like format data) were removed for model training and evaluation purposes. Additionally, to prevent the FM from learning highly individualized features, we removed  measurement categories that appeared in fewer than 200 patient records.

\paragraph{Tokenization details}
We tokenized our clinical data following a methodology similar to ETHOS, incorporating a comprehensive set of temporal tokens ranging from 1 day to 10 years to capture diverse time scales. Numerical measurement values were transformed into quantiles, with each quantile specific to its corresponding measurement, ensuring contextually relevant representations. Additionally, we explicitly inserted "start visit" (specific to each visit type) and "end visit" tokens to delineate clinical encounters within the health trajectories.

Initially, there were 630M medical events across all participants spanning 80k unique tokens in the OMOP format. After the filtering process, 150k participants  remain in our dataset, which present 20k unique tokens (codes and auxiliary tokens), yielding an average of 1500 tokens per participant. Health trajectories may span up to decades of data. We chose to work with the raw OMOP clinical events, rather than abstracting information, which leads us to having a relatively large vocabulary compared to extant work. The dataset, comprising data from AoU 150,000 participants, was randomly split at a 90:10 ratio into a training set (135,000 participants) and a held-out evaluation set (15,000 participants) with a splitting ratio of 90/10.

\paragraph{Genomics to PRS}
Summary statistics for a total of 7,145 GWAS were downloaded from the 20200615 snapshot of the Pan-UKB Reference Consortium data.\footnote{\href{https://pan.ukbb.broadinstitute.org}{https://pan.ukbb.broadinstitute.org}}
For each GWAS, “independent hits” were identified using the PLINK\footnote{\href{https://www.cog-genomics.org/plink/1.9 }{https://www.cog-genomics.org/plink/1.9 }} -- clump command ~\cite{purcell2007plink}. Lead variants were identified as having p-value $\leq 1 \times 10^{-4}$, with secondary variants clumped if they had p-value $\leq 1.1 \times 10^{-4}$ and $R^2 \geq 0.1$. The reference panel for linkage disequilibrium (LD) calculation contained 10,000 unrelated subjects of European ancestry from the UK Biobank.

Sample PRS values were computed for each sample, for each of the 7,145 GWASs, using the above “independent hits” as candidate variants. Only candidate variants with association p-value $\leq 5 \times 10^{-8}$ were retained.

Variants were mapped from GRCh37 to GRCh38 with a best-effort handling of cross-assembly strand and ref/alt changes (which cannot always be unambiguously resolved). After filtering and mapping, 3481 traits with at least one significant variant remained. For each trait, at most 1500 variants (ranked by absolute beta value) were selected per trait for computing the polygenic risk scores (PRS). The PRS was computed for each trait and participant as the sum of the product of beta values and number of effect alleles present. Before ingesting the PRS into our models, we clipped the values to remove outliers. The overall calculation pipeline is represented in Figure \ref{fig:prs-calculation}.

\subsection{Models and multimodality}

\begin{figure}[h!]
  \centering
  
  % Subfigure (a)
  \begin{subfigure}[b]{1.0\textwidth}
    \centering
    \includegraphics[width=\textwidth]{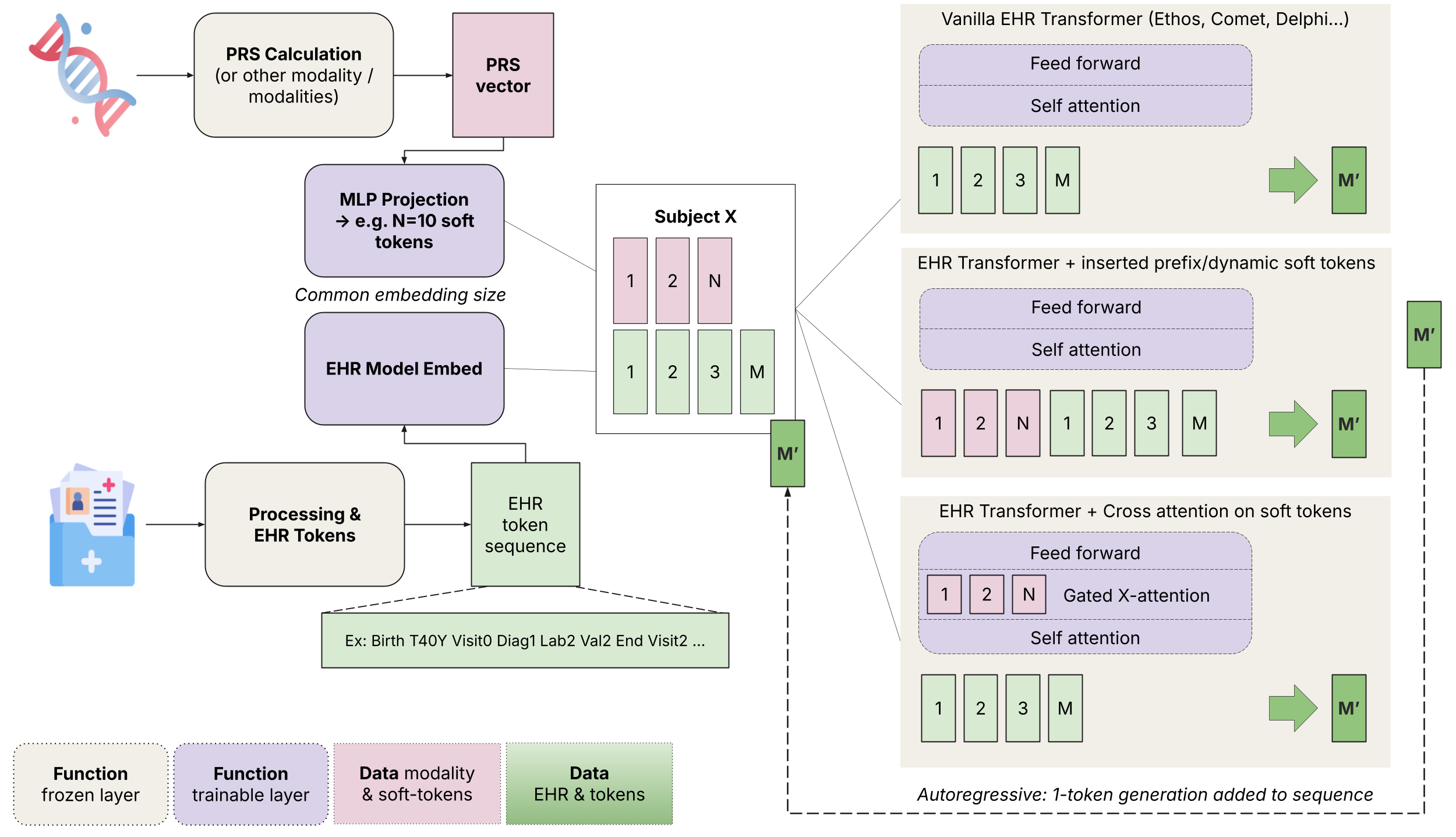}
    \caption{Proposed architectures for integrating PRS into EHR FM.}
    \label{fig:framework}
  \end{subfigure}
  
  % Horizontal line between subfigures
  \vspace{0.8em}
  \rule{0.8\textwidth}{0.5pt} % (width, thickness)
  \vspace{0.8em}
  
  % Subfigure (b)
  \begin{subfigure}[b]{1.0\textwidth}
    \centering
    \includegraphics[width=\textwidth]{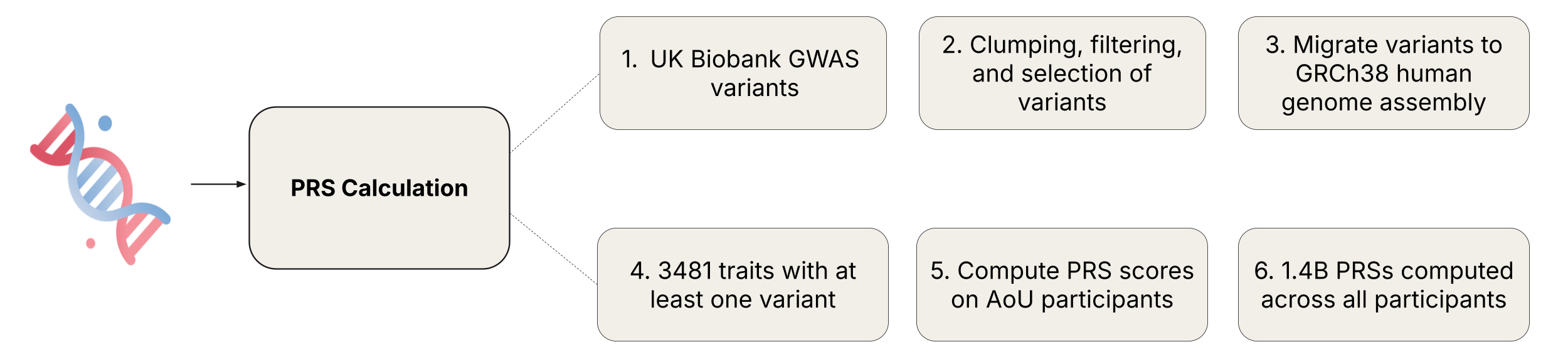}
    \caption{PRS Calculation pipeline.}
    \label{fig:prs-calculation}
  \end{subfigure}

  \caption{
  Proposed framework for integrating various data modalities, such as polygenic risk scores (PRS) into Electronic Health Records (EHR) foundational models. 
  \textbf{(top)} Architecture suggested: details how different inputs are processed, embedded, and then fed into Transformer architectures, including variations with inserted prefix/dynamic soft tokens and cross-attention mechanisms, to generate health trajectories (M') for downstream tasks. This work focuses on comparing the basic EHR only model against the transformer with cross-attention for PRS.
  \textbf{(bottom)} Data calculation pipeline from raw genomics GWAS to PRS on All of Us dataset.}
  \label{fig:two_parts_vertical}
\end{figure}

All models evaluated here were trained on a GPT-2 (hugging face version) decoder with cross-attention to incorporate external modalities into this transformer \cite{vaswani2017attention} based EHR FM.
The proposed cross-attention architecture extends to all models enabling cross attention. For other transformer architecture without cross-attention natively enabled, we propose the inserted soft tokens in future work as a natural extension mirroring advancements in multimodal LLMs. We summarize the framework in Figure \ref{fig:framework}.

For every version of the model, unless specified otherwise, we use the same backbone for the auto-regressive model: an embedding size of 960 for the approximately 20k tokens (~20M parameters), window size of 2048 which specifies the maximum context length including for generation, 12 layers and 8 model heads. This totals to a base model of 155 million parameters for the base decoder GPT-2 using EHR only.

All models are trained to maximize next token prediction likelihood, also understood as cross-entropy loss over the distribution of language model head. We use a training batch size of 2, with each entry corresponding to a single participant. We parallelize our training across 8 V100 GPUs available on the All of Us secure platform, yielding an effective batch size of 16. We used AdamW as our optimizer with a learning rate set to 1e-4 and a linear scheduler across our entire training pass. Models were all trained through 5 epochs of randomly shuffled data with no specific data sampling nor modality pacing.

\paragraph{Multimodal static features}
In the main part of this study, we focused on integrating the vectorized 3481 PRS into the EHR base model. Genomics, and PRS in particular, are modeled as static individual attributes that can affect the next token prediction at any stage. In this work, we focused on PRS to illustrate how to incorporate external modalities into EHR FM.

As mentioned above, we proposed 2 frameworks for integrating other modalities into our model: cross attention or inserting soft tokens. Regardless of the chosen framework (cross attention or inserted soft tokens), it is necessary to project the vector of PRS into the embedding space of the GPT model. To do so, we used a 2-layer fully connected MLP (multi-layer perceptron) with Gelu activation in the middle, and a scaled sigmoid head to bring all values between (-1,1). We used a hidden layer size of 4096 and output layer size of 960x10 to produce 10 soft tokens, matching the embedding space and adding an additional 20 million parameters to the model. These are trainable model parameters that are learned jointly with the GPT model weights.

For cross attention, the projected 10 tokens are tended through the additional cross attention layers within each model head. These add an additional 45M parameters, and can allow a flexible size of number of tokens through the cross attention. This approach can be naturally extended to multiple modalities which may be turned on or off.

The soft token approach prepends projected tokens to the embeddings of the encoded health trajectories, effectively augmenting the window. These soft tokens have no token meaning and cannot be decoded, but are used when computing the attention mechanism (self attention or cross attention), and influence the prediction of future tokens.

We summarize in Table~\ref{tab:multimodal_approaches}  different options to incorporate external modalities into EHR FM.

\renewcommand{\arraystretch}{1.3} % increases row height
\begin{table}[ht]
\centering
\caption{Comparison of different approaches for incorporating multimodal features into models.}
\label{tab:multimodal_approaches}
\begin{tabular}{p{3cm} p{2.5cm} p{4.5cm} p{4.5cm}}
\toprule
Approach & Feature type & Pro & Con \\
\midrule
\textbf{Soft token prefix} & Static (e.g., genomic, SDoH) & Medium difficulty. Easily use multiple modalities. Can easily augment data to drop modalities. & May be lost with long history. Not rich information. Utilizes tokens from EHR in window. \\
\textbf{Soft token positional} dynamically incorporated in sequence & Dynamic (e.g., wearable, clinical notes, image) & Can drop modalities in training for data augmentation. & High difficulty. At inference, cannot directly generate soft tokens, lose consistency. Would benefit training a reconstruction layer (e.g., diffusion / MAE, harder). \\
\textbf{Soft token in each head} & Static (if prefix) / Dynamic (if positional) & Each head holds different information. & Very-high difficulty. No obvious gain over utilizing more soft-tokens. \\
\textbf{Cross attention}, \textit{approach used in this paper} & Static & Relatively easy. Explicitly attend static modalities to every new query token. & Not enabled natively by all model architectures. Does not extend to dynamic modalities. Many more parameters. \\
\textbf{Ensemble of models} & Static / Dynamic & Easiest. & No joint learnings across different modalities. \\
\bottomrule
\end{tabular}
\end{table}

\paragraph{Naming}
In the remainder of the paper, we refer to the GPT-EHR or EHR model as the base model trained on EHR only data, and GPT-PRS-CROSS or EHR+PRS as the model trained using the 3481 PRS using cross attention.
Unless explicitly stated, we do not consider in depth the multimodal version with prepended soft tokens.

\paragraph{Note on dynamic modalities}
One salient feature of the soft token architecture representation is that it allows for the incorporation of the additional projected modalities at any point in the health trajectory. Although in our work we place the soft tokens as a prefix – and this is mostly reasonable given PRS genomics are static individual features – the soft tokens extracted from an imaging report or clinical notes can be positioned at the end of the related visit, i.e., dynamically within the health trajectory.

\paragraph{Positional embedding and training label note}
In our training we keep one training entry per participant. We insert left-padding on all training entries to the maximum window length. One could also consider concatenating training trajectories by separating them with end-of-sequence and start-of-sequence special tokens. This concatenation has been proven very efficient in pre-training for text-based LLMs \cite{renc2024zero}, but does not extend easily to the cross-attention framework.

As mentioned, pre-training optimizes the cross entropy loss for training tokens. In our experiments with prepended soft tokens, we excluded soft tokens from the loss computation as commonly done in multimodal decoders. There is no normalization across participants to up-sample participants with sparse EHR data, which may be an improvement to reduce bias.

For future work, in the case of dynamic modalities, one can consider "reconstructing" soft tokens (e.g., from wearables, clinical notes, imaging, etc.) from the EHR in order to keep the trajectories consistent. We leave as future work how best to incorporate dynamic modalities into the training with an appropriate reconstruction loss term.

\subsection{Extract tokens from multimodal data}
Another sophisticated approach to integrating multimodal data involves tokenizing the modalities with specific tokens. This method necessitates a deep understanding of the diverse data types and the extraction of highly relevant, interpretable features from each modality. For instance, natural language processing (NLP) techniques can be employed to extract critical information from clinical notes, such as symptoms, diagnoses, and treatment plans. Similarly, quantifiable metrics like high polygenic risk scores (PRS) can be directly incorporated as buckets or quantiles, providing insights into genetic predispositions. Specific survey questions, often designed to capture subjective individual experiences or lifestyle factors, can also be transformed into meaningful tokens. Furthermore, extracted features from radiology images, such as tumor characteristics or organ measurements, can be meticulously tokenized.
Once these diverse, interpretable features are extracted, they are then converted into a sequence of tokens, allowing for seamless integration into advanced analytical models. This methodical tokenization ensures that the rich and varied information from all modalities is represented and utilized in the downstream transformer.

While interpretability is preserved with this tokenization, this approach requires domain expertise to extract relevant features, and extensive manual supervision. This further anchors on known patterns, rather than reasoning about which additional information may be relevant to the transformer beyond what is already captured by the EHR. On the other hand, methods described above use the self-supervised learning to automatically identify relationships between the modalities and the EHR data. The fully self-supervised framework we propose does indeed offer a much more scalable method to integrate several modalities without requiring deep and across domain knowledge – which would otherwise hinder cross-domain learnings.

\subsection{A novel risk score for generated trajectories: the path-computing probabilities}
Although the generative model is pre-trained in a generic next-token-prediction fashion, EHR FMs have demonstrated their Zero-shot ability to predict various operational and clinical risks, through classification tasks e.g., “will an ER visit happen in the next 30 days?”.
In existing literature, for a given task, N different trajectories are drawn through Monte Carlo sampling, using the health trajectory up to time zero as context, where N is a parameter of the prediction task.

For each of the N generated trajectories, the algorithm inspects whether the target event (e.g., an ER admission) happens before the target time range has elapsed, by reading through the generated tokens. Then the empirical frequency of realization is utilized as a prediction score for the classification task, i.e., as an empirical estimate of:
$$\mathbb{P}[\text{target before time range} \mid \text{history before time zero}]$$

This flexible framework allows estimating the risk of various events that may involve multi-target sequences under flexible time horizons.
In our work, we only allow the model to generate K tokens where K is pre-specified for a given task, so we filter testing cases that have fewer than window size - K tokens. Another valid approach would allow generating more tokens while constantly left truncating the history and adapting the static tokens accordingly.

While the Monte Carlo sampling with empirical realization results in decent performing AUCs, we note that these models result in N+1 operating points, given that prediction scores take value in 0, 1/N,… 1.
Utilizing predictions for risk stratification may be hindered by these bucketed scores, as stratification typically requires the ability to identify finer grained risks, in order to select the “best” operating point on an ROC or precision-recall curve.

We have addressed this limitation using a novel path computing probability. Similarly, we first sample N Monte-Carlo trajectories from time zero for K new tokens. Along each path we cumulatively score the probability of a target token. More formally, using as context the tokens before time zero, and token$_i$ is understood as the $i$-th token along the generated trajectory $t$ truncated at the time range, we define the path computed probability estimate as:

\begin{align*}
    &\text{Estimator } \mathbb{P}[\text{target before time range on path }t] :=
    \sum_{\text{token}_i \in t} \prod_{j<i} (1-p_j(t)) \cdot p_i(t) \\
    &\text{Where } p_i(t) := \mathbb{P}[
        \text{token$_i$ = target} \mid 
        \text{history + token$_1$ + $\dots$ + token$_{i-1}$}
    ] 
\end{align*}

This probability is calculated through Bayesian update rules. Now each of the N paths has a path computed probability in [0,1]. These scores per path were aggregated (mean, median or max as specified) to get the estimated probability given the context. Unless specified otherwise, we use mean aggregation of path computed probabilities as the prediction score.
We have found these scores to be much more discretized than the original N+1 buckets from the Monte Carlo frequency estimate. Path computed probabilities yield roughly unique thresholds (understood as different operating points on the ROC curve) scaling with the number of test samples. Further these path computed probabilities generally present higher AUROC (Area Under / the Receiver Operating Characteristic Curve). Note this definition naturally extends to multiple target tokens.

To summarize, this novel computation enables us to get a much more nuanced understanding of individual risk as every prediction is roughly unique, and extracts more meaningful signal from the generated trajectories. In turn, this leads to continuous ROC curves and enables more granular operational decision making. When clear of context, we refer to AUROC as AUC.

\subsection{Comparative Evaluation of Model performance}
\paragraph{Model performance comparison via paired bootstrap}
Model performances of EHR-only model (Model A) and EHR+PRS model (Model B) were evaluated using threshold-free metrics (AUC and AUPRC), and thresholded metrics (accuracy, precision, recall, F1, and specificity, at some chosen threshold). The difference on each evaluation metric between models was estimated via paired bootstrap, and its mean estimate, 95\% confidence interval (CI) estimate, and two-sided p-value (with regard to null hypothesis of no difference between models) were reported. 
For each dataset, the disease labels and the corresponding model-predicted probabilities of each participant were read; if multiple scores per participant were present (typically for the path computed probabilities), the average of them were aggregated as the predicted probability .
Details of the paired bootstrap were as follows
\begin{itemize}
    \item In each bootstrap iteration, $N$ subjects (equal to the dataset size) were sampled with replacement, the metric difference between Model A and Model B is computed as $\Delta$ = metric(B) - metric(A). Positive $\Delta$ indicates Model B outperforms Model A.
    \item This process was repeated for I total iterations, and the mean delta, its 95\% CI (i.e., 2.5th and 97.5th percentiles of the bootstrapped distribution), and the two-sided p-value (i.e., estimated as the proportions of negative deltas and positive deltas) were computed.
\end{itemize}

% We compared two classifiers (model A vs. model B) on the same subjects using a paired, nonparametric bootstrap. For each dataset, we read per-subject labels y and model-predicted probabilities; if multiple scores per subject were present (typically for the path computed probabilities), they were aggregated (mean, median or max as specified). We evaluated threshold-free metrics - AUROC and AUPRC (Area Under the Precision Recall Curve), together with thresholded metrics (accuracy, precision, recall, F1 and specificity) at a fixed decision threshold (0.5). For each metric, we estimated the differences $\Delta$ = metric(B) - metric(A) via paired resampling: in each iterations, we sampled n subjects (equal to the dataset size) with replacement, preserving the pairing of A and B predictions within each resample, computed the metric for A and B, and stored the delta. We report the bootstrap mean delta, 95\% percentile confidence intervals (CI) (2.5th - 97.5th percentiles of the delta distribution), and a two-sided p-value based on the proportion of positive vs. negative bootstrap deltas (ties split evenly), which tests $H_0: \: \Delta=0$. Positive $\Delta$ indicates Model B outperforms Model A.

\paragraph{Precision–Recall Difference Analysis}
To compare classifiers across the full range of operating points, we evaluated differences in precision at fixed levels of recall. Predicted probabilities from Models A and B were first aligned at the subject level, with multiple predictions per subject aggregated using the mean. Precision–recall (PR) curves were constructed for each model, and precision values were interpolated on a fixed recall grid spanning 0 to 1.

At each recall level, we computed the difference in precision between the two models, $\Delta$(recall) = Precision$_B$(recall) $-$ Precision$_A$(recall).
To quantify uncertainty, we performed paired nonparametric bootstrap resampling. In each iteration, subjects were resampled with replacement, preserving the alignment of predictions from both models. Precision differences were recalculated on the recall grid; resamples containing only one outcome class were excluded. For each recall point, we report the bootstrap mean difference and the 95\% percentile interval (2.5th - 97.5th percentiles). Results were visualized as mean difference curves with shaded confidence intervals, where positive values indicate a precision advantage for Model B over Model A at the corresponding recall level.

\paragraph{Meta-analysis across prediction tasks}
To summarize model comparisons across multiple independent prediction tasks, we conducted a meta-analysis of paired performance differences. For each task, the paired bootstrap procedure (described above) produced an estimated effect size, defined as the difference $\Delta$ in a given metric between Model B and Model A (for example $\Delta$AUROC or other metric, at a fixed threshold), along with its bootstrap standard error.

Per-task estimates were pooled using inverse-variance weighting under both fixed-effect and random-effects models. The fixed-effect model assumes that all tasks estimate the same underlying effect, and weights studies according to the inverse of their variance. The random-effects model, implemented via the DerSimonian–Laird method, incorporates an additional between-task variance component ($\tau^2$) to account for heterogeneity.

Between-task heterogeneity was quantified using Cochran’s Q statistic, its associated $\chi^2$ test, and the I$^2$ statistic, which expresses the proportion of total variability attributable to heterogeneity rather than sampling error. Following conventional guidance, fixed-effect results were emphasized in the presence of low heterogeneity (I$^2$ $\leq$ 25\% or Q p $\geq$ 0.10), while random-effects estimates were prioritized when heterogeneity was moderate to high. In the case of few available tasks (k $\leq$ 5), we favored random-effects results due to the limited power of Q. The meta-analysis yields a pooled estimate of the effect size, its standard error, 95\% confidence interval, and a two-sided p-value for testing the null hypothesis of no difference between models.

\section{Evaluation and Results}

\subsection{Test set loss calculation}
We calculated the test set loss for all models on a held-out test split of over 15,000 participants (10\% of overall cohort), which were not used during self-supervised training or model calibration. To ensure a fair comparison, we evaluated both the model with no genomic GPT-EHR and the model employing cross-attention GPT-PRS-CROSS. For statistical validation, we performed a paired statistical test across participants in our test set.
We first assessed the differences in loss values between the two models using a t-test under normality assumption. In case the differences were found to be non-normal, we applied the Wilcoxon signed-rank test. Both tests consistently yielded statistically significant p-values (p < .001 for t-test and p < .05 for Wilcoxon test), indicating a robust difference in test loss performance.
All statistical tests were performed using the python \texttt{scipy.stats}\footnote{\href{https://docs.scipy.org/doc/scipy/tutorial/stats.html}{https://docs.scipy.org/doc/scipy/tutorial/stats.html}}
package.

The test loss serves as a direct extension of the self-supervised loss function, yet only acts as a surrogate for overall model usefulness. However we contend that it is a clear indicator of the model's goodness of fit onto the unseen test data. A lower test loss signifies that the model more accurately predicts future tokens in unseen health trajectories, implying a better understanding of underlying health dynamics.
In the remainder, we will pivot our focus to a more comprehensive evaluation, encompassing both qualitative assessments for deeper interpretability and quantitative analyses for predicting various clinically and operationally relevant events of interest.

\subsection{Event prediction cohort curation}
Each evaluation cohort was curated from the held out test split (unseen participants during self-supervised training and model calibration), for which specific labels were curated.
Cohorts are specifically designed to test predictions that are either general clinical outcomes, or selected to showcase the value of the modality. 

\subsubsection{Condition prediction from first time token}
We investigated the predictive power of our multimodal foundation model for 10-year disease onset, using only an individual's initial demographic data and their PRS. This task begins at "time zero" (the start of the participant's EHR), from which the model generates synthetic future health trajectories to forecast the appearance of specific disease codes. This minimal-data setting is designed to isolate the predictive value of the PRS modality.

Our primary analysis focuses on type 2 diabetes (T2D). The list of target tokens for T2D is compiled by selecting the OMOP concept ids associated with concept names containing the lowercase stem “diab” and “type 2” in any order. We then intersect the resulting 95 target tokens with those kept by our tokenizer’s vocabulary, resulting in 32 distinct target disease tokens. The final test cohort includes 12,009 participants with available PRS, of whom 1,720 developed T2D within the 10-year window. For each participant, we generated 10 synthetic trajectories via Monte Carlo sampling to predict disease onset.

\begin{figure}[ht]
    \centering
    \includegraphics[width=0.7\linewidth]{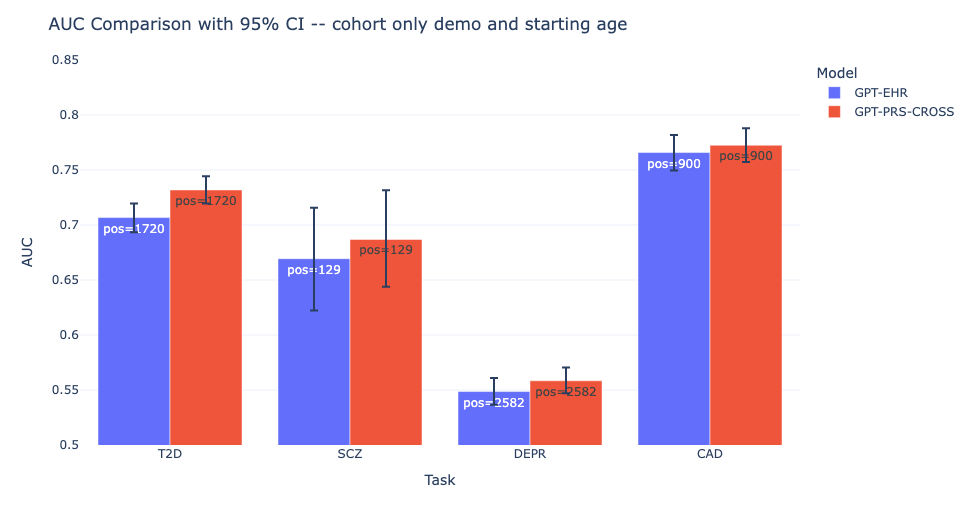}
    \caption{AUC Comparison between GPT-EHR and GPT-PRS-CROSS, on 10 year prediction with demographic tokens only, for various conditions: type 2 diabetes (T2D), schizophrenia (SCZ), depression (DEPR), and coronary artery disease (CAD). 
    Positive number of cases for each task is shown. Test size = 12,009. T2D has non-overlapping bootstrapped confidence intervals. The multimodal model consistently improves over the EHR only baseline.}
    \label{fig:auc-conditions}
\end{figure}

As shown in Figure \ref{fig:auc-conditions} and Table \ref{tab:t2d_stats}, the EHR+PRS model combining demographic tokens (gender, sex, and race) and PRS significantly outperformed the EHR-only model with demographic tokens for T2D prediction. We observed statistically significant improvements across all metrics, including a 15.5\% relative increase in AUPRC ($\Delta$=0.041, p<0.0001) and a 3.6\% increase in AUROC ($\Delta$=0.025, p<0.0001). A precision-recall difference analysis confirmed that the multimodal model improves precision across most of the recall range, suggesting the value of PRS is most pronounced for individuals at higher genomic risk.

For other conditions tested - coronary artery disease (CAD), depression, and schizophrenia (SCZ) - the addition of PRS showed directionally-consistent gains but did not yield statistically significant improvements (Figure \ref{fig:auc-conditions}).

\begin{table}[ht]
\centering
\begin{tabular}{lcccccc}
\toprule
Metric & A:EHR only & B:EHR+PRS & $\Delta$ (B $-$ A) & 95\% CI for $\Delta$ & p-value & \% Change \\
\midrule
AUPRC & 0.269 & 0.311 & +0.041 & [0.026, 0.056] & <0.0001 & +15.5\% \\
AUROC & 0.707 & 0.732 & +0.025 & [0.016, 0.034] & <0.0001 & +3.6\% \\
Accuracy @0.50 & 0.843 & 0.851 & +0.007 & [0.003, 0.011] & 0.0005 & +0.9\% \\
Precision @0.50 & 0.352 & 0.433 & +0.080 & [0.036, 0.123] & 0.0010 & +23.0\% \\
Recall @0.50 & 0.113 & 0.140 & +0.026 & [0.007, 0.044] & 0.0050 & +23.7\% \\
F1 @0.50 & 0.171 & 0.211 & +0.040 & [0.013, 0.065] & 0.0010 & +23.5\% \\
Specificity @0.50 & 0.965 & 0.969 & +0.004 & [0.001, 0.008] & 0.0190 & +0.4\% \\
\bottomrule
\end{tabular}
\caption{
Model performance comparison between Model A (EHR only) and Model B (EHR+PRS). Metric differences were estimated using paired bootstrap with 2,000 iterations, with metric difference defined as $\Delta$=metric(B)-metric(A). Two-sided p-value (with regard to null hypothesis of no difference between models) are reported.
}
\label{tab:t2d_stats}
\end{table}

\begin{figure}
    \centering
    \includegraphics[width=0.5\linewidth]{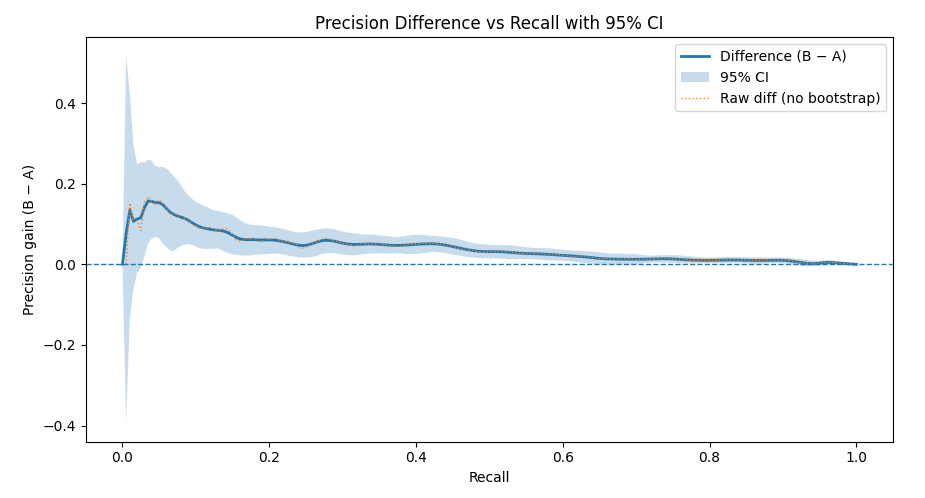}
    \caption{Precision-Recall Curve difference between model EHR+PRS (B) and model with EHR only (A) for 10-year T2D prediction, using demographic tokens.}
    \label{fig:t2d-pr}
\end{figure}

\paragraph{Meta-analysis of $\Delta$AUROC across tasks}
We combined $\Delta$AUROC estimates from the four disease prediction tasks (T2D, CAD, DEPR, SCZ) using both fixed- and random-effects models. Individual task-level differences ranged from +0.007 to +0.025. Heterogeneity was moderate (Q = 7.63, df = 3, p = 0.054; I$^2$ = 60.7\%), suggesting variability in effect sizes across tasks.
Under the fixed-effect model, the pooled $\Delta$AUROC was +0.016 (95\% CI: 0.010–0.022, p < 0.0001). The random-effects model, which accounts for between-task heterogeneity, produced a slightly attenuated but still significant pooled estimate of +0.015 (95\% CI: 0.004–0.026, p < 0.01). These findings indicate that incorporating PRS into EHR-based models yields a modest but statistically significant improvement in discrimination across diseases, with the strongest evidence in T2D and consistent positive trends in other tasks.

\begin{figure}
    \centering
    \includegraphics[width=0.5\linewidth]{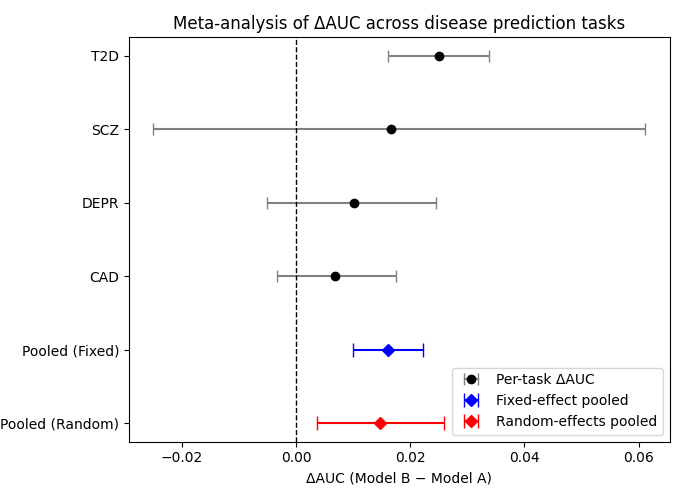}
    \caption{Meta-analysis of $\Delta$AUROC across tasks}
    \label{fig:meta}
\end{figure}

\subsubsection{Influence of longer EHR history on predictive power of PRS on T2D onset prediction }
To explore how EHR history length affects the predictive power of PRS for type 2 diabetes (T2D) onset, we established a "time zero" three years after each participant's earliest EHR data. We created four distinct observation windows: 0 months, 6 months, 1 year, and 3 years of EHR history. To align time zero, we inserted time tokens between demographics and EHR data.

We excluded participants with a T2D condition token before time zero and those with less than five years of EHR data, resulting in a cohort of 7,802 members with 679 positive labels. The model, using an established vocabulary of clinical events, predicted T2D onset by generating N=10 distinct sequences, each K=512 tokens long, via Monte Carlo sampling.

Our hypothesis was that the predictive power of PRS would decrease as more EHR history was incorporated. This is because a richer EHR history is expected to capture an increasing amount of clinical information relevant to T2D onset, thereby reducing the incremental value of static genetic predispositions. The results supported this: the incremental benefit of PRS, measured by $\Delta$AUROC, was largest with 0 months of EHR ($\Delta$ = 0.0223, 95\% CI 0.0064–0.0371, p < 0.005). This benefit attenuated at 6 months ($\Delta$ = 0.0114, p = 0.14) and 1 year ($\Delta$ = 0.0108, p = 0.087), becoming statistically insignificant at 3 years ($\Delta$ = -0.0024, p = 0.70).

\begin{figure}[ht]
    \centering
    \includegraphics[width=0.7\linewidth]{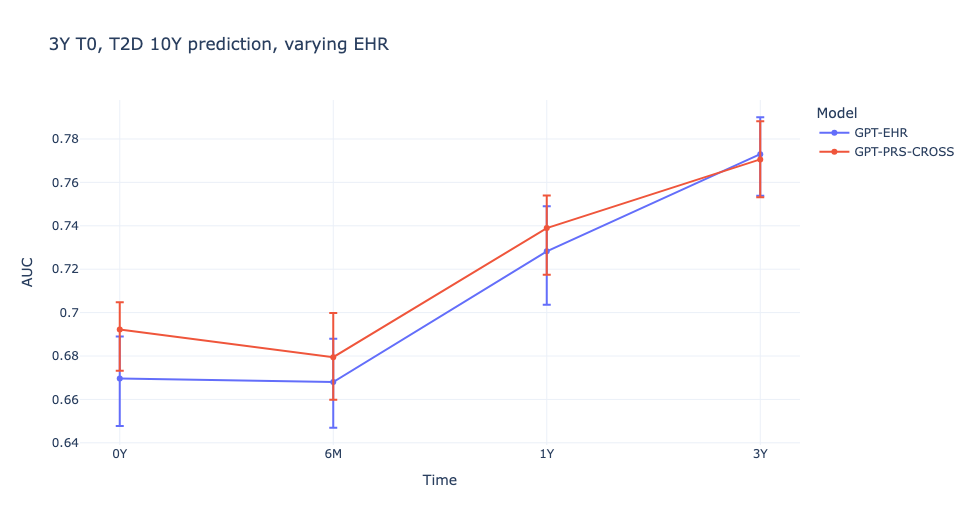}
    \caption{AUC Comparison between GPT-EHR and GPT-PRS-CROSS, on T2D onset 10 year prediction with varying amount of EHR history prior to time zero (0, 6 months, 1 year, 3 years). Test size = 7082.
    More EHR history leads to more accurate predictions.
    Gains of the multimodal model relative to the EHR baseline fade as an increasing amount of EHR is made available.
    }
    \label{fig:varying-history}
\end{figure}

\subsubsection{Effect of path computed probability}
\begin{figure}[ht]
    \centering
    \includegraphics[width=0.7\linewidth]{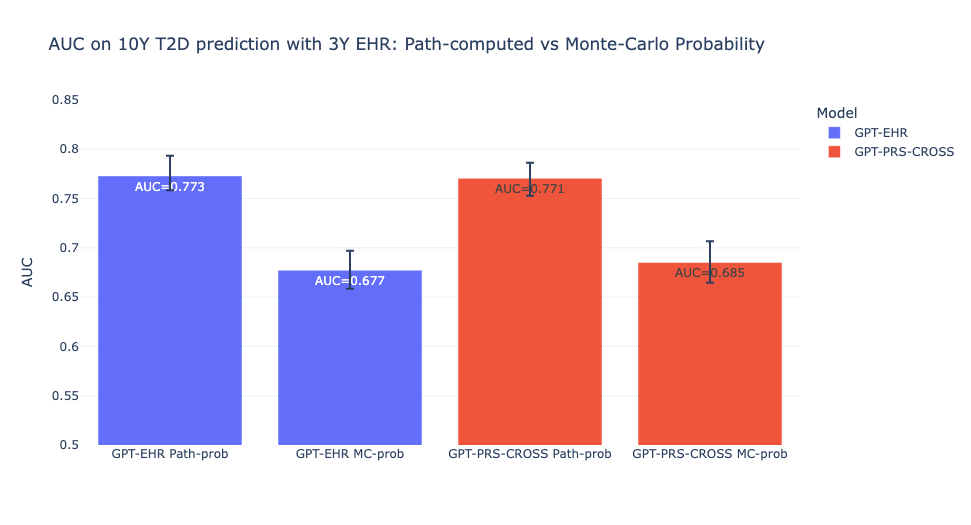}
    \caption{AUC Comparison between Monte Carlo estimates and Path computed probabilities for GPT-EHR and GPT-PRS-CROSS, on T2D onset 10 year prediction with 3 years of EHR. Test size = 7082.
    Path computed probabilities lead to significantly higher AUC.
    }
    \label{fig:probas-bars}
\end{figure}

To illustrate the effect of using path computed probabilities, we anchor on the 10 year T2D prediction above, using 3 years of EHR in the history. We compare side by side the effect of using the raw Monte Carlo empirical frequencies vs our proposed prediction scores.
It is quite clear that for both models, there are substantial and significant gains when using path computed probabilities (Figure~\ref{fig:probas-bars}).

\begin{figure}[ht]
    \centering
    \includegraphics[width=0.7\linewidth]{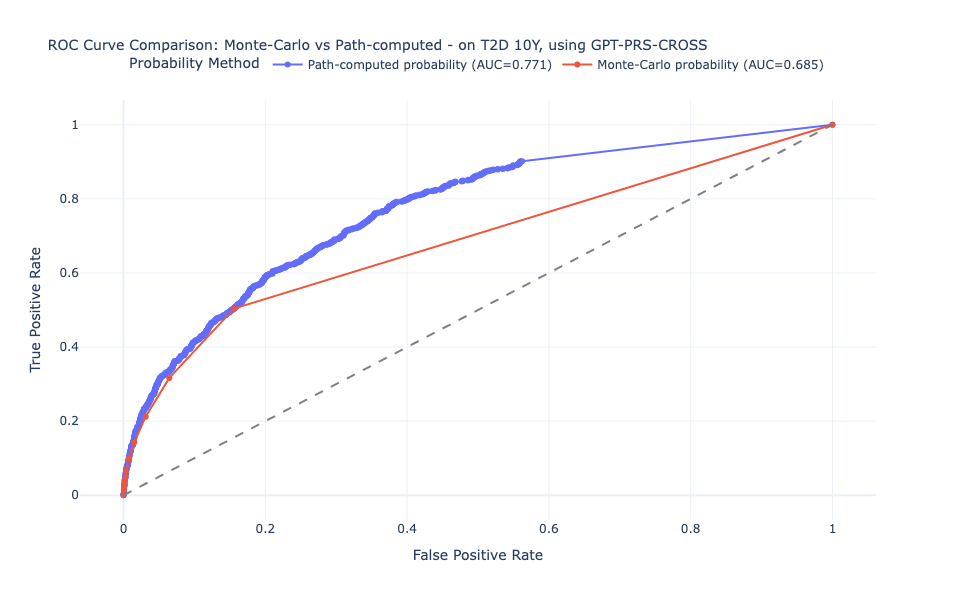}
    \caption{ROC Comparison between Monte Carlo estimates and path computed probabilities for GPT-PRS-CROSS, on T2D onset 10 year prediction with 3 years of EHR. Test size = 7082.
    Path computed probabilities leads to a higher AUC, and a much more granular choice of operating points, compared to the 10 enabled by Monte Carlo estimates.
    }
    \label{fig:probas-auc}
\end{figure}

This novel probability computation offers a significant advantage by enabling the selection of a more precise operating point from an ROC curve. As depicted in Figure~\ref{fig:probas-auc}, the probabilities derived from this path computation provide a substantially greater number of operating points compared to the mere 10 (equal to the number of samples) observed with Monte Carlo samples. This advancement has the potential to facilitate more granular clinical and operational decision-making, by directly making the generated health trajectories more relevant to individual risk stratification.

\subsection{Correlation between model prediction and known PRS}
To understand how the inclusion of polygenic risk scores (PRS) impacts model predictions for specific conditions, we conducted an analysis focusing on a specific task, type 2 diabetes (T2D) predictions using the cohort with no EHR history, as described above.

We identified known PRS associated with the condition, leveraging established genetic literature and the comprehensive set of 3481 PRS derived from the Pan-UKB Reference Consortium data. For each participant in the test cohort, we calculated the difference between the T2D risk prediction probabilities generated by the GPT-PRS-CROSS model and the GPT-EHR (base) model. This difference, denoted as 'probability-differences', quantifies the incremental predictive contribution of integrating PRS into the multimodal FM compared to the base model.

Subsequently, we computed the Spearman rank correlation between these 'probability-differences' values and the identified T2D-specific PRS for each participant. Our hypothesis was that a positive correlation would be observed indicating that the GPT-PRS-CROSS model's predictions are directionally influenced by key PRSs associated with the condition of interest. In turn, this would highlight the value of the self-supervised learning process, even across modalities. We show only p-values for positive Spearman rank correlation in the tables calculated using \texttt{scipy.stats}.

% TODO Pouya / Alessandra
\begin{table}[ht]
\centering
\caption{Spearman Rank positive correlation p-values between probabilities-differences for T2D using no EHR. * indicates that the target PRS itself is significantly positively correlated with the actual T2D diagnosis in the cohort (p<.05). Only p-values are shown.}
\begin{tabular}{lc}
\toprule
Target & p-value for positive correlation \\
\midrule
Label defined T2D 10Y * & 2.6003e-09 \\
Self-reported type 2 diabetes (categorical-20002-both\_sexes-1223) * & 1.1494e-03 \\
Doctor-diagnosed diabetes (categorical-2443-both\_sexes-2443) * & 7.9553e-08 \\
Self-reported diabetes (categorical-20002-both\_sexes-1220) * & 5.1003e-07 \\
\bottomrule
\end{tabular}
\label{tab:t2d_prs}
\end{table}

Table \ref{tab:t2d_prs} displays the p-values from a Spearman rank correlation analysis for the T2D prediction task using no prior EHR data. It illustrates that GPT-PRS-CROSS learns signals associated with each of these known PRS, which are indeed associated with the T2D label. Furthermore, as a control, we also ensured that the known T2D-associated PRS themselves exhibited a significant positive correlation with the actual T2D label in our cohort.

Our analytical framework was expanded to include coronary artery disease (CAD) and its established PRS in Table \ref{tab:cad_prs}. The results consistently demonstrate that PRS exhibiting a significant positive correlation with the CAD label (indicated by an asterisk for statistical significance) show 'probability-differences' that align with these correlations. This alignment indicates that the predictive capacity of these specific PRS for CAD is consistently reflected in the calculated probability shifts from EHR-GPT to EHR-PRS-CROSS.

\begin{table}[ht]
\centering
\caption{Spearman Rank positive correlation p-values between probabilities-differences for CAD using no EHR. * indicates that the target PRS itself is significantly positively correlated with the actual CAD diagnosis in the cohort (p<.05). Only p-values are shown.}
\begin{tabular}{lc}
\toprule
Target & p-value for positive correlation \\
\midrule
Label defined CAD 10Y * & 4.4717e-03 \\
Self-reported myocardial infarction (categorical-20002-both\_sexes-1075) * & 9.5536e-03 \\
No doctor-diagnosed heart problems (categorical-6150-both\_sexes-100) & 9.9998e-01 \\
Doctor-diagnosed heart attack (categorical-6150-both\_sexes-1) * & 1.4611e-02 \\
ICD10 I24 code "Other acute..." (icd10-I24-both\_sexes) & 9.9018e-01 \\
ICD10 I25 code "chronic ischemic heart disease" (icd10-I25-both\_sexes) * & 3.8438e-01 \\
ICD10 I51 code "Complications and..." (icd10-I51-both\_sexes) & 9.9581e-01 \\
\bottomrule
\end{tabular}
\label{tab:cad_prs}
\end{table}

This analysis validates the underlying genetic risk factors' relevance to the disease outcome, thereby reinforcing the interpretability of any observed correlations with our multimodal model's predictions. The results of this analysis provide crucial insights into how our multimodal framework effectively leverages genomic information to refine risk predictions, particularly in scenarios where traditional clinical data is limited.

\subsection{Transfer learning experiments}
As an alternative to sampling full trajectories to predict disease occurrence, we also show it is possible to transfer the foundation model for custom classification tasks. We explored two methods of transfer learning: direct fine-tuning and PRS feature extraction.

\subsubsection{Example of fine-tuning with 10-year stroke prediction}
We chose 10-year disease onset for stroke, defined broadly as the inclusion of any OMOP code corresponding to cerebrovascular accident, cerebral hemorrhage, or cerebral infarction. The prediction window was indexed from a random visit in the participant’s history.

We added a linear layer with dropout (p=0.1) on top of the frozen multimodal transformer to perform a direct binary classification, where the pooled representation from the last non-padded token position is projected from the transformer's hidden dimension (960) to the number of output classes (2).
\begin{table}[ht]
\centering
\caption{Comparison of model inference complexity and predictive performance for 10-year stroke prediction (transfer learning). $K$ = \# tokens to generate, $N$ = \# Monte Carlo generated trajectories}
\begin{tabular}{lccc}
\toprule
Model & Inference Complexity & AUROC & AUPRC \\
\midrule
Demographics (logistic) & $O(1)$ & 0.51 (0.49--0.52) & 0.08 \\
GPT (auto-regressive) & $O(K \cdot N)$ & 0.61 (0.59--0.63) & 0.14 \\
GPT (fine-tuned classifier) & $O(1)$ & 0.64 (0.62--0.66) & 0.14 \\
\bottomrule
\end{tabular}
\end{table}

The resulting fine tuned model performs comparably to the base model with Monte Carlo sampling, and is much more efficient for classification because it only requires a single forward pass to produce classification logits rather than auto-regressive token-by-token generation. Eliminating the computational overhead of sequential decoding is useful for discriminative tasks like disease onset where only the final classification probability is needed.

\subsubsection{Example of feature transfer with 10-year COPD}
We chose 10-year disease onset for Chronic Obstructive Pulmonary Disease, defined by 13 target OMOP codes. The prediction window was indexed from birth, so no EHR data is included as input.

For each individual, we performed a single forward pass and mean-pooled the output of the genomics sub-model to generate an embedding of size 960. We then trained a lightweight MLP classifier with one hidden layer with 100 neurons on top of these embeddings. Prior to fitting, embeddings were standardized (per feature, using training statistics).

We compared the performance of this model with that of a similar model using raw PRS scores (size 3481) instead of their learned projected embeddings, as well as a demographics baseline. Models were evaluated for statistical significance using the same test splits and participant-pairwise bootstrap method described above.

\begin{figure}[h!]
  \centering

  \begin{subfigure}[b]{.45\textwidth}
    \centering
    \begin{tabular}{lcc}
        \toprule
        Alternate & p-value & p-value \\
        hypothesis  & wrt demographics & wrt PRS-raw \\
        \midrule
        $\Delta$AUROC $> 0$ & 0.05 & 0.04 \\
        $\Delta$AUPRC $> 0$ & 0.11 & 0.07 \\
        \bottomrule
    \end{tabular}
    \label{tab:bla}
    \caption{Model performance comparison between MLP with embedded PRS and baseline models for 10-year COPD prediction. Two-sided p-value (with regard to null hypothesis of no difference comparing embedded-PRS model with baseline) model reported.}
  \end{subfigure}
  \hfill
  \begin{subfigure}[t]{.45\textwidth}
    \centering
    \includegraphics[width=1.\linewidth]{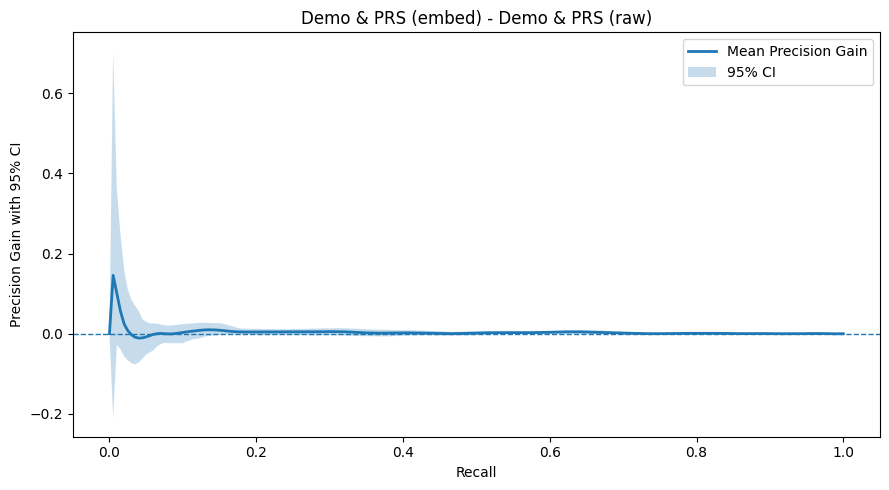}
    \caption{Precision Recall difference between MLP using embedded PRS vs. raw PRS, on 10-year COPD prediction (transfer learning), no EHR is included in the input.
    }
    \label{fig:transfer-pr}
  \end{subfigure}
  \caption{Feature transfer experiments on COPD.}
  \label{fig:transfer_feature_copd}
\end{figure}

The model using embedded PRS features demonstrated a marked increase in precision at the low-recall end of the curve. This indicates that for the task of identifying the cohort of individuals at the highest risk, our model is more reliable, yielding fewer false positives than the model using raw features.
Results and p-values are shown in Figure \ref{fig:transfer_feature_copd}.

\section{Discussion}
\subsection{Summary of Results}
Our work introduces an innovative Electronic Health Record (EHR) foundation model (FM) that incorporates polygenic risk scores (PRS), moving beyond single-modality EHR approaches to create a more holistic health profile. We trained this novel multimodal model on the diverse data from the All of Us (AoU) Research Program, leveraging its rich EHR and genomic data to develop deeper insights.

We adapted modern generative AI techniques to integrate these external modalities, allowing the model to learn complex relationships between clinical data and genetic predispositions. This approach is key to unlocking new insights for disease prediction, risk stratification, and personalized treatment strategies.

Our evaluation on AoU data was designed to show the model's comprehensive utility. While we shouldn't expect the multimodal nature of our model to have clinical value across all prediction tasks, we can reasonably compare our model with EHR only FM for all benchmarks, and still highlight significant gains in areas related to the incorporated modalities. We first assessed the direct impact of PRS integration on early disease detection for conditions with known genetic links: type 2 diabetes (T2D), schizophrenia (SCZ), coronary artery disease (CAD), and chronic obstructive pulmonary disease (COPD).

Next, we analyzed how the model weighs genetic predispositions against accumulated clinical history to refine predictions as more EHR data becomes available. We also validated that our model's predictions align with established PRS, confirming it learned meaningful biological signals.

Finally, we demonstrated the versatility of our architecture by using its learned encoders for transfer learning tasks. We adapted the pre-trained multimodal representations to new, specific downstream prediction problems, assessing their performance in fine-tuning and feature extraction scenarios.

% \paragraph{Value to Product and Patients}
Our model extends beyond simple predictions by generating a spectrum of plausible future health trajectories for an individual, allowing providers to explore "what-if" scenarios. It performs dynamic risk assessment, continuously updating risk profiles as new data becomes available.

A crucial capability is the creation of high-fidelity synthetic patient/participant data, which mirrors real data but contains no private information. This addresses data scarcity for rare diseases, enhances privacy, and allows researchers to experiment before using sensitive real-world datasets.

Ultimately, our vision is a "digital twin" \cite{Makarov2025}—a high-fidelity virtual replica of an individual's health. This digital twin could simulate the progression of conditions and test therapeutic strategies virtually, providing continuous, personalized recommendations for predictive and preventative healthcare.

\subsection{Limitations and Future Work}
\paragraph{On Data and Bias}
There are also several limitations that need to be considered when interpreting the findings of this study.
First, the specific cohort utilized from the AoU Research Program introduces certain limitations. The demographic and health characteristics of the AoU cohort, while diverse, may not perfectly represent the broader population, potentially affecting the generalizability of our findings. Furthermore, the reliance on pre-existing diagnostic codes for identifying conditions within the EHR data presents inherent challenges. These codes can sometimes be incomplete, inaccurate, or inconsistently applied across different healthcare systems, which may lead to mislabeling of diseases or conditions, thus impacting the accuracy of our analyses.

Second, issues encountered during the filtering and construction of the study cohort warrant attention. Filtering rules based on lengths of medical codes were applied to the overall AoU cohort, which was necessary for maintaining data quality and taking into account model complexity (for practical model training purposes). This process reduced the sample size (relative to the overall AoU cohort) and introduced some selection bias, by excluding participants who may have had a greater burden of disease.  Further work lies ahead to fully characterize the effect that this selection may have had on the representativeness of our cohorts as it relates to intended target populations. Future efforts will focus on refining these filtering strategies to maximize the utility of the AoU data while minimizing potential biases.

Third, this study primarily focused on PRS and their applications. Future research would benefit from incorporating other valuable modalities of data available within the AoU program, such as detailed survey data, objective measurements from wearable devices, high-resolution medical images, and rich clinical notes. Jointly evaluating modalities and how to train across several modalities is unclear. Addressing these challenges is crucial for a more comprehensive and individualized understanding of health outcomes.

% Several limitations must be considered. First, the AoU cohort, while diverse, may not be perfectly representative of the broader population, which could affect the generalizability of our findings. Second, our reliance on pre-existing diagnostic codes in EHR data can lead to inaccuracies and misclassification.
% Third, our study focused primarily on PRS. Future work should incorporate other AoU data modalities like survey data, wearable device measurements, medical images, and clinical notes. Integrating these diverse data types presents significant computational challenges, including data harmonization and developing appropriate multimodal analytical frameworks.

% Finally we applied inclusion and exclusion criteria, necessary for maintaining data quality and analytical focus (for practical model training purposes). This process reduced our sample size (relative to the overall AoU cohort) and introduced some selection bias. Further work lies ahead to fully characterize the effect that this selection may have had on the representativeness of our cohorts as it relates to intended target populations. Future efforts will also focus on refining filtering and training strategies to maximize the utility of the AoU data (and other datasets) while minimizing potential biases.

\paragraph{On PRS Limitation and Applicability}
% The utility of PRS is subject to limitations. The generalizability of PRS derived from predominantly European ancestry cohorts remains a concern, which may reduce predictive accuracy for non-European populations in the AoU cohort. A direct mitigation strategy would be to up-sample underrepresented ancestries during model training and evaluation.

% Additionally, PRS is just one method of incorporating genetic information. Future research could model specific genetic variants directly, using more sophisticated architectures to interpret rare, high-effect variants or capture non-additive interactions. Our experiments show that the predictive power of PRS can diminish as an individual accumulates more comprehensive EHR data. This highlights the need for models that can dynamically weigh the influence of various modalities over time.

The utility of PRS within the AoU dataset, while significant, is subject to certain limitations that warrant careful consideration. Primarily, the generalizability of PRS derived from predominantly European ancestry cohorts remains a concern. While AoU aims for diverse representation, many existing GWAS that inform PRS development have historically been biased towards individuals of European descent. This can lead to reduced predictive accuracy when applying these PRS to non-European populations within the AoU cohort, potentially exacerbating health disparities. A direct attempt to mitigate bias would be to up-sample under-represented PRS or ancestries within model training and evaluation.

It is also important to recognize that PRS are just one way of incorporating genetic information into a predictive model. While they provide a valuable summary of an individual's genetic predisposition, they condense complex genetic architecture into a single score per trait. Other approaches, such as modeling specific genetic variants directly, represent promising future avenues of research. This could involve using more sophisticated model architectures to interpret the impact of rare, high-effect variants or to capture non-additive interactions between common variants that are not fully represented by a linear PRS. Such methods could provide a more granular and potentially more accurate representation of genetic risk.

Finally, the dynamic nature of health means that genetic risk, while static, interacts with an individual's evolving health trajectory. The predictive power of PRS might diminish or be modulated as a participant accumulates more comprehensive EHR data over time, as demonstrated in our experiments with increasing EHR history. This highlights the need for models that can dynamically weigh the influence of various modalities as more information becomes available, rather than treating PRS as a universally constant predictor.

\paragraph{Limitations in Training and Evaluation}
Our study used GPT-2 as the foundational architecture, with a focus on cross-attention for static modalities. Future work will explore more contemporary transformer models with longer context windows, for example built on Mamba architecture \cite{wornow2024context}, using Rotary Position Embeddings (RoPE), and other state of the art attention mechanisms etc. Further research could incorporate parametric encoding of time as in \cite{shmatko2025learning}, and more robust hyperparameter optimization. We will also transition from a single-sequence-per-line training approach to one that concatenates multiple health trajectories. Data augmentation techniques, such as strategically omitting visits or progressively introducing new modalities, are also critical areas for future research.

Within the EHR FM community, evaluating the quality of generated trajectories beyond conventional metrics is unclear. We believe evaluation should be guided by specific, practical use cases to ensure outputs are not only statistically sound but also clinically meaningful.

We did not assess the calibration of our model, thereby limiting the direct clinical utility of this research. While our GPT-PRS-CROSS model demonstrated improved discrimination, as evidenced by the change in AUC, confirming that the model can effectively rank patients by risk, we did not analyze model calibration. Calibration measures whether the predicted probabilities correspond to the actual observed event rates.  Our primary focus was on the novel architecture and the proof-of-concept demonstration that integrating PRS improves discriminative performance, making AUC the most direct metric for this goal. To bridge the gap between the model’s demonstrated discriminative performance and its potential for clinical application, future work should incorporate a comprehensive calibration  analysis to confirm the predicted absolute risk scores accurately correspond to observed event frequencies. This analysis will ensure the reliability of the model’s risk predictions so that it can be considered for use in patient care. 

% \paragraph{On Prediction and Sampling}
Our primary goal is to show the feasibility and utility of integrating external data modalities into EHR foundation models, not to achieve state-of-the-art performance on every task. The main strength of our approach is its principled framework for multimodal integration and risk score estimation, which enhances the diversity and interpretability for FMs. Future work can further optimize predictive power and achieve benchmark-leading performance. 

Post-training techniques such as supervised fine-tuning, which we briefly explored, and reinforcement learning (RL) are known to improve performance on downstream tasks. Similar to how reasoning models can be trained with verifiable rewards, our model could be rewarded to produce more relevant outputs for a given task based on prediction accuracy. This iterative feedback loop would guide the model toward more accurate predictions. As with RL in text-based LLMs, we would cautiously balance tuning for the specific problem while preserving key properties of the pre-trained foundation model for interpretability.
Another area for future exploration is ensembling our model's zero-shot predictions with task-specific state-of-the-art methods. This hybrid approach would combine the strengths of our generalizable foundation model with highly specialized predictive algorithms, potentially leading to further performance gains.

Ideally, a foundational model should have a strong capacity for few-shot learning, enabling it to be effective even with limited examples. This capability is crucial for rapid deployment and adaptation in new domains where extensive labeled datasets are scarce, maximizing the model's utility across a wide range of applications. We believe these future directions will move the field beyond foundational EHR models and enable the creation of truly intelligent, personalized, and preventative healthcare solutions.

\section{Acknowledgments}
This project represents an extensive collaboration between numerous teams at Verily Life Sciences and NVIDIA. We express our gratitude to Mohammad Azimi, Youpeng Su, Nicholas Bense, Shanhong Chen, Yu Hu, June Hua, David Shen, and Alessandro Culotti from Verily, as well as Gary Burnett from NVIDIA, for their instrumental contributions in enhancing the Verily Pre platform with GPU-accelerated libraries and frameworks, including Parabricks and NeMo, and modern GPUs (H200, B200), thereby enabling efficient pre-training.
We further acknowledge Moira Dillon, David Glazer, Suki Singh, Myoung Cha, Bharat Rajagopal, Scott Burke, Ilia Tulchinsky, and Michael Radwin from Verily, alongside Trent Norris and Kimberly Powell from NVIDIA, for their efforts in navigating cross-company boundaries, facilitating data usage, and ultimately fostering this collaboration.

We gratefully acknowledge All of Us participants for their contributions, without whom this research would not have been possible. We also thank the National Institutes of Health’s All of Us Research Program, including the team at the Data and Research Center, for making available the participant data examined in this study.

% \bibliographystyle{ieeetr}
% \bibliography{references}

\end{document}